# *CLAIMED*
# the open source framework for building coarse-grained operators for accelerated discovery in science

**Romeo Kienzler[1], Rafflesia Khan[1], Jerome Nilmeier[2], Ivan Nesic[3], Ibrahim Haddad[4]**

[1]IBM Research Europe, [2]Bloomberg LP, [3]University Hospital Basel, [4]Linux Foundation AI

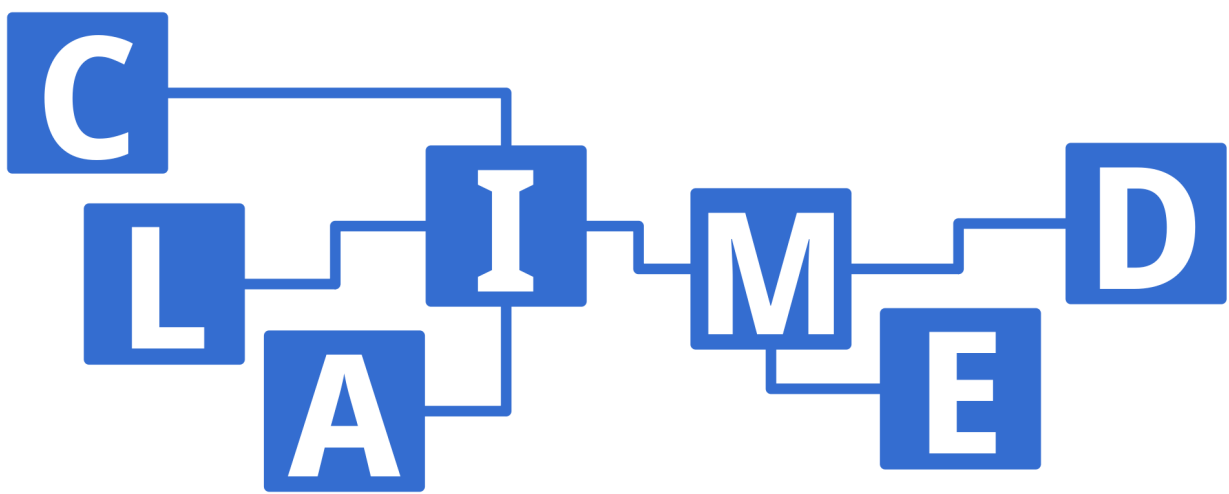

## *Abstract*

*In modern data-driven science, reproducibility and reusability are key challenges. Scientists are well skilled in the process from data to publication. Although some publication channels require source code and data to be made accessible, rerunning and verifying experiments is usually hard due to a lack of standards. Therefore, reusing existing scientific data processing code from state-of-the-art research is hard as well. This is why we introduce CLAIMED, which has a proven track record in scientific research for addressing the repeatability and reusability issues in modern data-driven science. CLAIMED is a framework to build reusable operators and scalable scientific workflows by supporting the scientist to draw from previous work by re-composing workflows from existing libraries of coarse-grained scientific operators. Although various implementations exist, CLAIMED is programming language, scientific library, and execution environment agnostic.*

# Introduction

In our different Research Departments, we frequently collaborate with domain subject matter experts, medical doctors, atmospheric physicists, and quantitative analysts (quants), for example, which, in this work, we call Citizen Data Scientists (CDS), who work extensively with large datasets in fields like computer vision, time series analysis, and NLP. However, we often encounter challenges with the existing approaches, such as using monolithic scripts for prototyping that lack quality and reproducibility. Therefore, in close cooperation with CDS, we have identified the following requirements to enhance the reusability and reproducibility of data-driven research:

- A low-code/no-code environment with visual editing and Jupyter notebooks for rapid prototyping
- Seamless scalability during development and deployment stages
- GPU support for efficient processing of big data
- Pre-built components tailored to various research domains
- Comprehensive support for popular libraries in the Python and R ecosystems, including Apache Spark, Ray, DASK, TensorFlow, PyTorch, pandas, and scikit-learn
- Easy extensibility to incorporate future advancements and techniques
- Ensuring the reproducibility of research work for reliable and transparent results
- Maintaining data lineage to track the origin and transformations of datasets
- Facilitating collaboration among researchers to foster knowledge sharing and teamwork.

We have evaluated various software tools commonly used in data science, HPC, and research, such as Slurm, Snakemake, QSub, HTCondor, Apache Nifi, NodeRED, KNIME, Galaxy, Reana, WEKA, Rabix, Nextflow, OpenWDL, CWL, and Cromwell. However, we found that none of these tools, even when combined, fully meet our requirements. To avoid reinventing the wheel, we have created the CLAIMED (Component Library for Artificial Intelligence, Machine Learning, Extract, Transform, Load, and Data Science) framework, a set of tools to create extensible operator libraries designed specifically for low-code/no-code environments in data-driven science. In the following section, we provide detailed insights into the implementation, followed by exemplary workflows that demonstrate the capabilities of CLAIMED within the context of climate research and life science. Furthermore, we discuss potential avenues for improving CLAIMED in the "Future Work" section, concluding with a summary of our findings.

# Implementation

CLAIMED is an operator library designed to create low-code and no-code workflows for applications in artificial intelligence, machine learning, ETL, and data science. This library provides ready-made components for various business domains, supports multiple computer languages, and runs on various execution engines, including Kubernetes, KNative Serverless, Kubeflow, Airflow, watsonx.ai [watsonxai] (MCAD) [mcad] and plain Docker. Our latest contribution is a command line tool that allows the usage of CLAIMED components interactively from the shell or in shell scripts. It's a notebook-based and visually stylized library, with a "write once, runs anywhere" approach. The primary goal of CLAIMED is to enable low-code/no-code rapid prototyping for seamless integration of CI/CD into production. CLAIMED's components can be run as Docker containers or Kubernetes services, and they can be exported for use with various engines. Considering all its features, IBM Watson Studio Pipelines includes the library as a first-class citizen.

# The CLAIMED Framework

CLAIMED is composed of three core components:

- The CLAIMED operator source code is a growing list of scripts and notebooks which resemble open source coarse-grained operators which are opinionated, tested and ready for production use
- The CLAIMED component compiler (C³) which takes any list of source code like the library mentioned above, creates deployable operators out of it and adds them to appropriate operator catalogs
- The CLAIMED operator consumers like the CLAIMED CLI, Kubeflow or IBM watsonx.ai

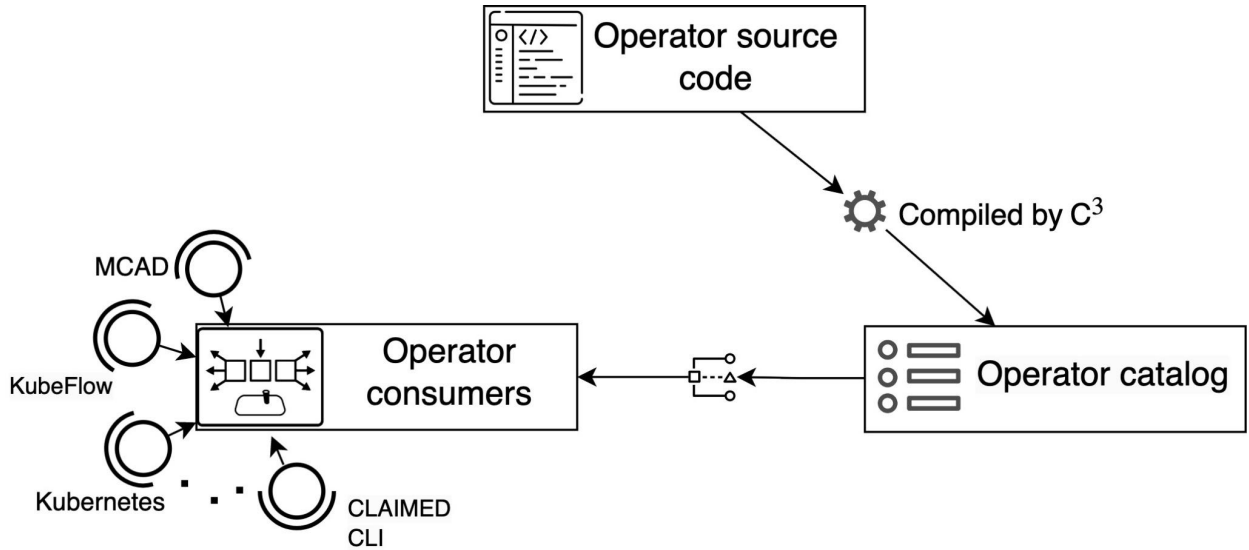

*Figure 1: How CLAIMED core components interact*

## C³ - The CLAIMED component compiler

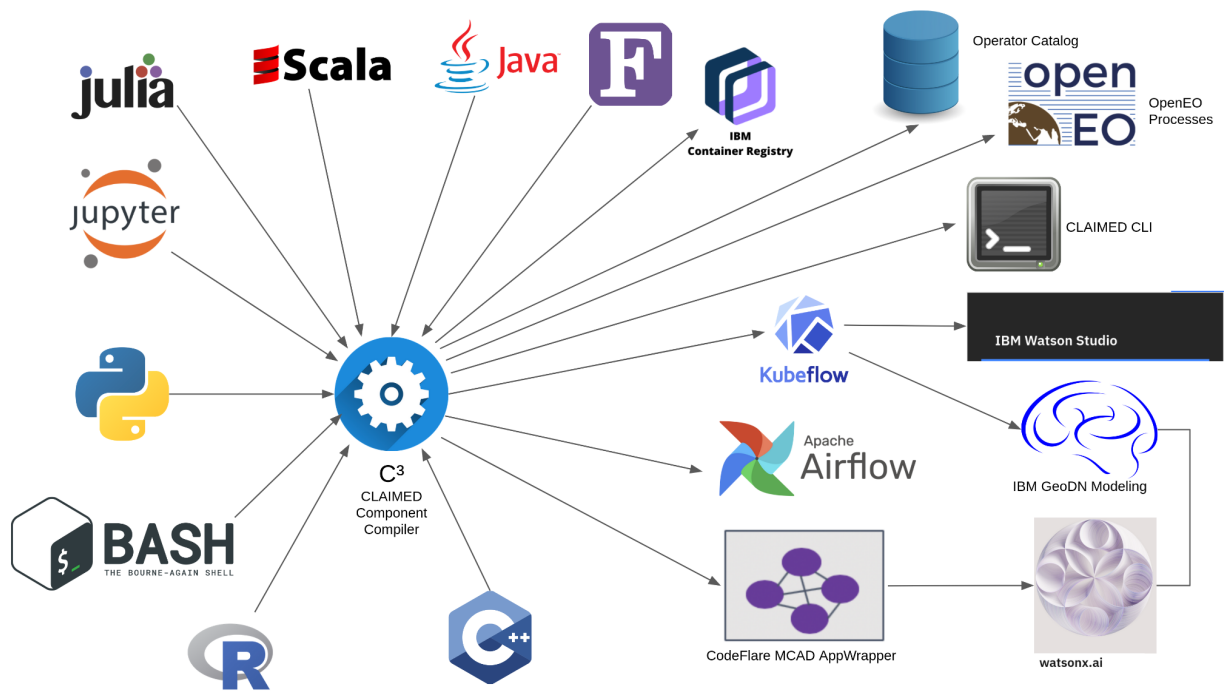

*Figure 2: C³ is language and framework agnostic. It automatically creates container images containing all code and required libraries in the correct versions, then pushes to a registry. In addition, C³ creates all necessary deployment descriptors for the different target runtime platforms.*

C³ is a compilation tool that accepts source code in different formats and programming languages and turns them into reusable operators. The key property of C³ is to let the operator's author focus on the development only. All additional requirements like creating Dockerfiles and Kubeflow Pipeline yaml deployment descriptors, for example, are taken care of by C³. The operator's interface is derived from the source code. The author must follow a minimal set of conventions to allow C³ to derive all necessary information for compilation (design by contract). If those conventions are not followed, C³ will inform the author with a detailed compilation error. We will exemplify C³'s function based on a jupyter python notebook with Kubeflow as the target runtime execution platform.

**1)** output-upload-to-cos

**2)** Uploads a file to any S3 compliant Cloud Object Storage

**3)**
```
!pip install aiobotocore botocore s3fs
```

**4)**
```python
import logging
import os
import sys
import re
import s3fs
```

**5)**
```python
# access key id
access_key_id = os.environ.get('access_key_id')

# secret access key
secret_access_key = os.environ.get('secret_access_key')

# cos/s3 endpoint
endpoint = os.environ.get('endpoint')

# cos bucket name
bucket_name = os.environ.get('bucket_name')

# source file to be uploaded
source_file = os.environ.get('source_file')

# destination file name
destination_file = os.environ.get('destination_file')

# temporary data folder
data_dir = os.environ.get('data_dir', '../../data/')

# dummy_output (to be fixed once C3 supports < 1 outputs)
output_dummy = os.environ.get('output_dummy', 'output_dummy')
```

**6)**
```python
parameters = list(
    map(lambda s: re.sub('$', '"', s),
        map(
            lambda s: s.replace('=', '="'),
            filter(
                lambda s: s.find('=') > -1 and bool(re.match(r'[A-Za-z0-9_]*=[.\/A-Za-z0-9]*', s)),
                sys.argv
            )
        )))

for parameter in parameters:
    logging.warning('Parameter: ' + parameter)
    exec(parameter)
```

**7)**
```python
s3 = s3fs.S3FileSystem(
    anon=False,
    key=access_key_id,
    secret=secret_access_key,
    client_kwargs={'endpoint_url': endpoint}
)
```

**8)**
```python
s3.put(data_dir + source_file, bucket_name + '/' + destination_file)
```

*Figure 3: Example of a python notebook based operator definition, by convention, containing all necessary information for C³ to create the Dockerfile, container image and Kubeflow Pipeline Component deployment descriptor (yaml)*

Let's have a look at Figure 3 to understand the conventions the developer needs to follow. Per definition:

- The first cell (1) contains the unique name of the operator (using canonical notation to indicate the category, e.g., 'output')
- The second cell (2) contains a description of the operator
- The third cell (3)installs all dependencies, e.g., using pip3. (interestingly, this code is able to run standalone as well as giving C³ enough information to install all library dependencies into the container image it creates)
- The fourth cell (4) imports all dependencies, imports are also allowed after this cell by C³ but it is considered as bad practice and linting tools like flake8 fail on default
- The fifth cell (5) contains all information for C³ to create proper operator interface definitions as all input and output parameters and types including default values and corresponding descriptions as comments are present in this cell
- The sixth cell (6) will be redundant and auto generated by C³ in the near future, but it allows, in addition to environment variables, to also access the operator's interface via command line parameters

Essentially, there are four mandatory cells which need to follow C³'s convention to give it enough information to create operator's for arbitrary target platforms. The two remaining cells, (7) and (8), in this example contain the operator's functional code - in this case - uploading a data file to cloud object storage.

It is important to note that C³ is implemented modularly. For example, the function to extract an operator's interface is a specific parser class extending from a generic interface and, in the case of Python source code, uses regular expressions only, as illustrated in Figure 4.

```python
class PythonScriptParser(ScriptParser):
    def search_expressions(self) -> Dict[str, List]:
        # TODO: add more key:list-of-regex pairs to parse for additional resources
        regex_dict = dict()

        # First regex matches envvar assignments of form os.environ["name"] = value w or w/o value provided
        # Second regex matches envvar assignments that use os.getenv("name", "value") with ow w/o default provided
        # Third regex matches envvar assignments that use os.environ.get("name", "value") with or w/o default provided
        # Both name and value are captured if possible
        envs = [r"os\.environ\[[\"']([a-zA-Z_]+[A-Za-z0-9_]*)[\"']\](?:\s*=(?:\s*[\"'](.[^\"']*)?[\"'])?)*",
                r"os\.getenv\([\"']([a-zA-Z_]+[A-Za-z0-9_]*)[\"'](?:\s*\,\s*[\"'](.[^\"']*)?[\"'])?",
                r"os\.environ\.get\([\"']([a-zA-Z_]+[A-Za-z0-9_]*)[\"'](?:\s*\,(?:\s*[\"'](.[^\"']*)?[\"'])?)*"]
        regex_dict["env_vars"] = envs
        return regex_dict
```

*Figure 4: C³ is a modular framework. This code illustrates the specific implementation for extracting the operator's interface - in this case - using regular expressions*

# CLI - The CLAIMED command line interface

For rapid prototyping on small datasets an extensive runtime environment like Kubeflow adds unnecessary overhead and complexity. Therefore, we've introduced the CLAIMED CLI - a command line interface that allows all CLAIMED operators to be used from the command line in the form of:

claimed <component_name:version> <parameters, ...>

For example, to list the contents of a COS bucket one would write:

claimed claimed-util-cos:0.3 access_key_id=xxx secret_access_key=yyy endpoint=https://s3.us-east.cloud-object-storage.appdomain.cloud bucket_name=era5-cropscape-zarr path=/ recursive=True operation=ls

As illustrated in Figure 5, when invoking a CLAIMED operator using the CLAIMED CLI the corresponding container image is pulled from the registry. These images have been created by C³ before. Then it runs the image using a local docker [docker] or podman [podman] installation. To allow for the sequential execution of subsequent operators, volumes are created dynamically for data transfer. For those operators supporting fully incremental data processing, data exchanges happen via TCP sockets using the HTTP protocol.

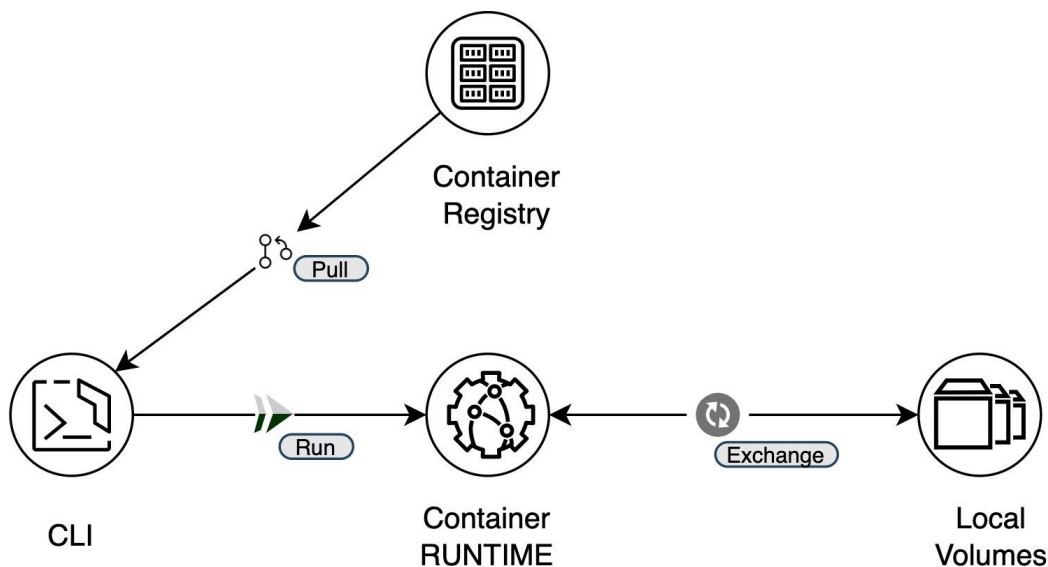

*Figure 5: The CLAIMED CLI pulls CLAIMED operator images from the container registry and runs them using a local container runtime environment like docker or podman. To enable data passing between subsequent executions of different CLAIMED operators local volumes are mounted into the containers*

# The open source CLAIMED operator reference codebase

Although emphasizing open source driven open science, sometimes it is necessary to keep operators private for legal or business strategy reasons. Therefore the CLAIMED ecosystem supports the idea of using the same tool suite for the public, open source operator code bases as well as proprietary licensed ones, allowing enterprises to create value add on top of open source CLAIMED for their internal and external clients as well as partner organizations.

Therefore, there is even more a need for an open source operator reference codebase serving as the foundation set of generic operators as well as a rich source of reference implementations illustration solutions to common problems.

As of June 2023, illustrated in Table 1, the open source operator base consists of 64 operators in 19 categories, which are available at [claimed]:

| Category Name | # of Operators |
| --- | --- |
| analyze | 1 |
| anomaly | 2 |
| checkpoint | 2 |
| deploy | 4 |
| filter | 2 |
| geo | 1 |
| input | 14 |
| metric | 3 |
| monitoring | 1 |
| nlp | 1 |
| output | 4 |
| predict | 5 |
| sim | 1 |
| train | 3 |
| transform | 14 |
| util | 2 |
| visualize | 2 |

*Table 1: As of June 2023 there exist 64 open source operators in 19 categories*

One important feature to notice is CI/CD support. Every change, once committed to git, triggers a rebuild and minor version increase on the operator level. Apart from releases, minor changes are tracked, and workflows can specify, or pin, exact operator versions allowing for reproducibility and audit.

# Example Use Cases

## Classification of Computer Tomography (CT) scans for COVID-19

With CLAIMED, users can easily create workflows from operators and run them on different platforms, ensuring a smooth and efficient developer experience. To demonstrate the utility of CLAIMED, a workflow was constructed using only the library's operators to classify Computer Tomography (CT) scans as either COVID-19 positive or negative. The publicly available [covidata] dataset was used to create a deep learning model for this purpose. The pipeline was built using Elyra's [elyra] Pipeline Visual Editor, which supports local, Airflow, and Kubeflow execution. To further emphasize the capabilities of CLAIMED, the AIX360/LIME [aix360] library was used to highlight the shortcomings of a poor deep learning model that only focused on the bones in the CT scans. Using CLAIMED, anyone without any coding experience can use the library operators in a low-code environment, drag-drop-connect, set their required parameters, and create proficient outcomes within no time.

**Figure:** CLAIMED use case (COVID-19 case detection using CLAIMED)

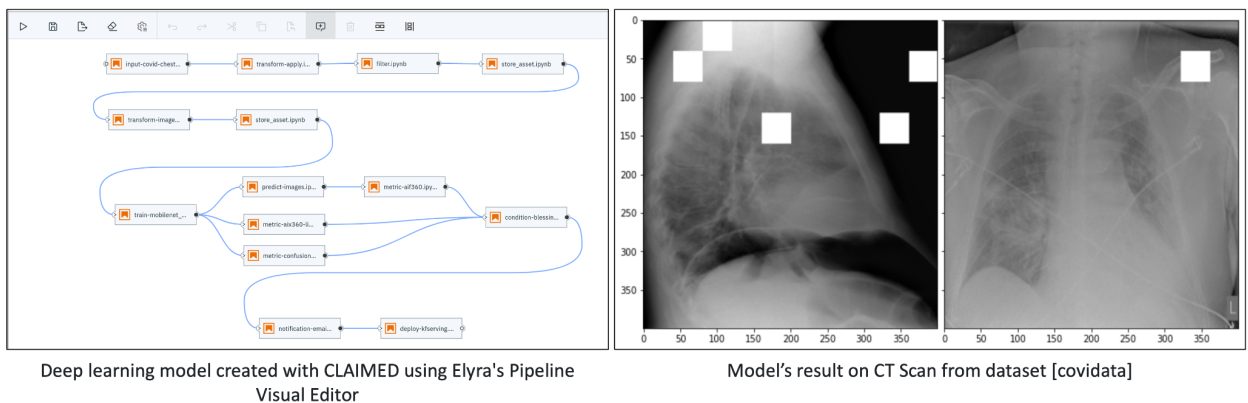

Deep learning model created with CLAIMED using Elyra's Pipeline Visual Editor

Model's result on CT Scan from dataset [covidata]

*Figure 6: Thanks to the already available AIX operators, the researchers at University Hospital Basel were able to identify a biased model. The model used latent information (gender, in this case, derived from bone structures) to improve classification performance as classes were heavily skewed on the gender dimension*

## Geospatial-Temporal Data Analysis Use Case

In geospatial-temporal data analysis, two main data source types exist. Earth observation (EO) data and climate data. EO data is mainly produced by satellites orbiting in non-stationary orbits (e.g., Sun-synchronous [sunsyn] for Sentinel-2), planes and drones, creating images of the planet in different temporal, spatial and radio spectral band resolutions. Climate data is usually measured or simulated geospatial-temporal data in different spatial and temporal resolutions created by different climate models or model ensembles.

To illustrate CLAIMED capabilities in the field of Geospatial-Temporal Data Analysis, we've created a set of reusable operators which were composed into a foundation model fine-tuning workflow, using one of IBM's geospatial foundation models [ibmfm], and pushed it to Kubeflow, provided by RedHat OpenShift Data Science [rhosds], for execution.

Let's have a look at Figure 7 to understand this workflow. In foundation model fine-tuning, which is related to transfer learning [tl], a generic, pre-trained model is used to learn a specific task. In this use case, we train a model to detect different Open Street Map (OSM) [osm] object types like houses, streets, and rivers to create a model to automate OSM annotations of unannotated regions.

First, we are querying and filtering data from Sentinel-2 [sentinel2] and Landsat [landsat] data sources using the **query-geodn-discovery** operator. Once brought to the same Coordinate Reference System (CRS) [crs] using the **regrid** operator we inner join these data on their spatial and temporal keys. In parallel, we pull vector data from OSM residing in a PostGIS database using the **postgis-connector** operators. The **generate-annotations** operator transforms the vector data into polygon - label pairs which are spatially joined with the satellite imagery data using the **join-spatial** operator. Images containing clouds are removed by the **mask-clouds** operator and images get normalized using the **normalize** operator to account for variations in lighting condition and color rendering. Now data is ready for the fine-tuning task. The **prithvi-finetune** operator is an example of how predefined operators can be combined with custom code. Here, the IBM geospatial foundation model *prithvi* is used as a head to a trainable U-net architecture. Once trained with training data derived from the **train-test-split** operator the model is tested using the test operator. If the metrics meet expectations, the model is automatically onboarded to the inference service of watsonx.ai [watsonxai] backed by Model Mesh [modelmesh].

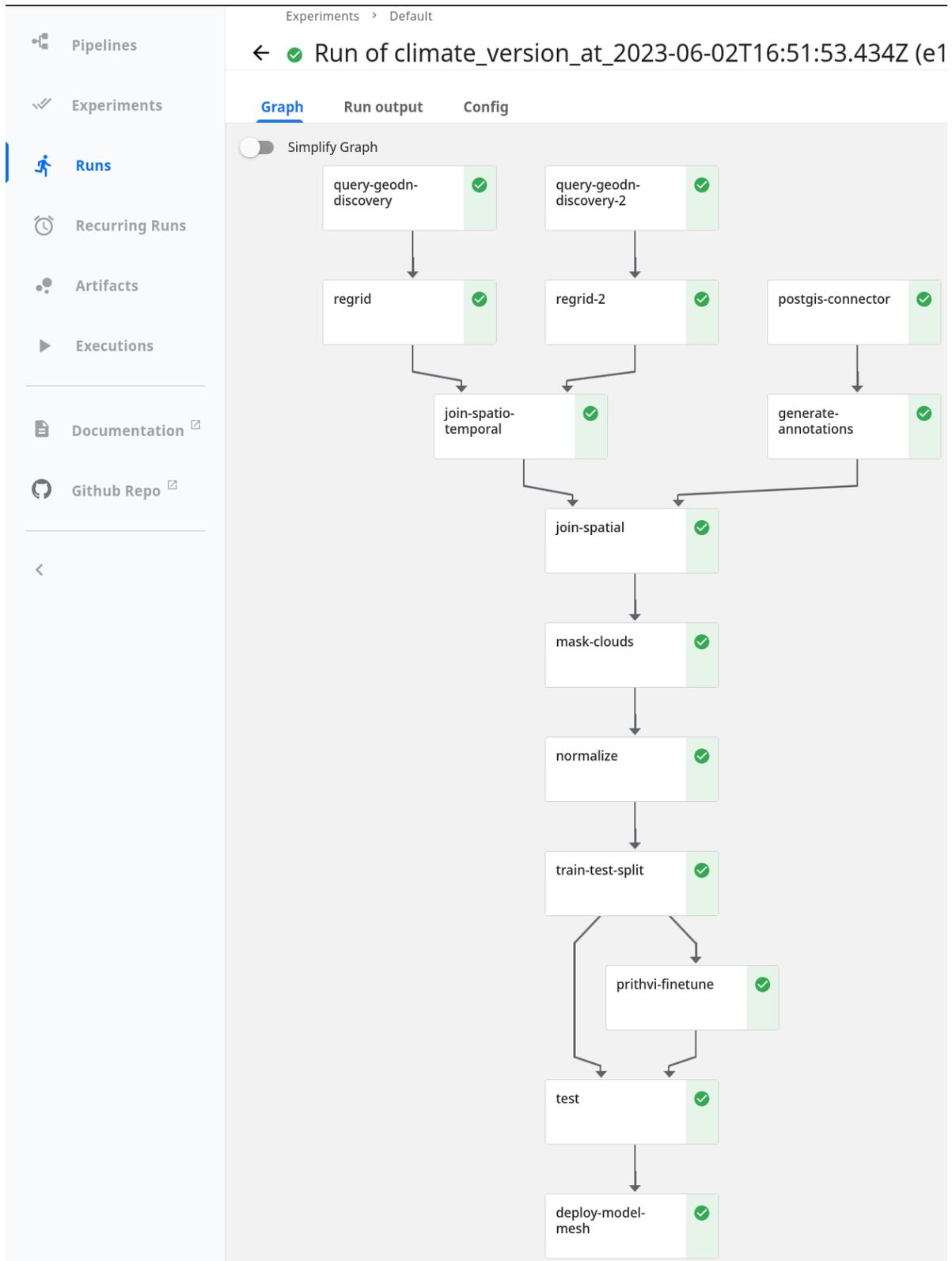

Figure 7 - The Geospatial-temporal workflow composed of CLAIMED operators executed on Kubeflow Pipelines

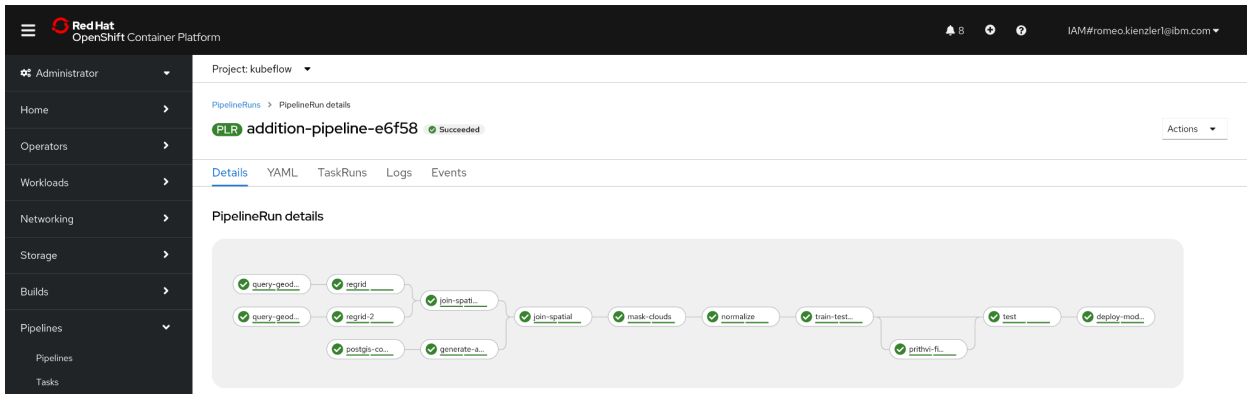

*Figure 8: Taking advantage of the tight integration of Kubeflow Pipelines into the RedHat OpenShift Data Science Platform*

# Project Analytics

On November 17, 2022, CLAIMED was accepted for Incubation at the Linux Foundation AI. As a benefit to hosting with the Foundation, CLAIMED is onboarded on the LFX Platform and as a result, we are able to have access to detailed project development analytics. The following charts offer various statistics of the projects and all of them were captured on May 30, 2023, and represent the last 3 years of development of CLAIMED.

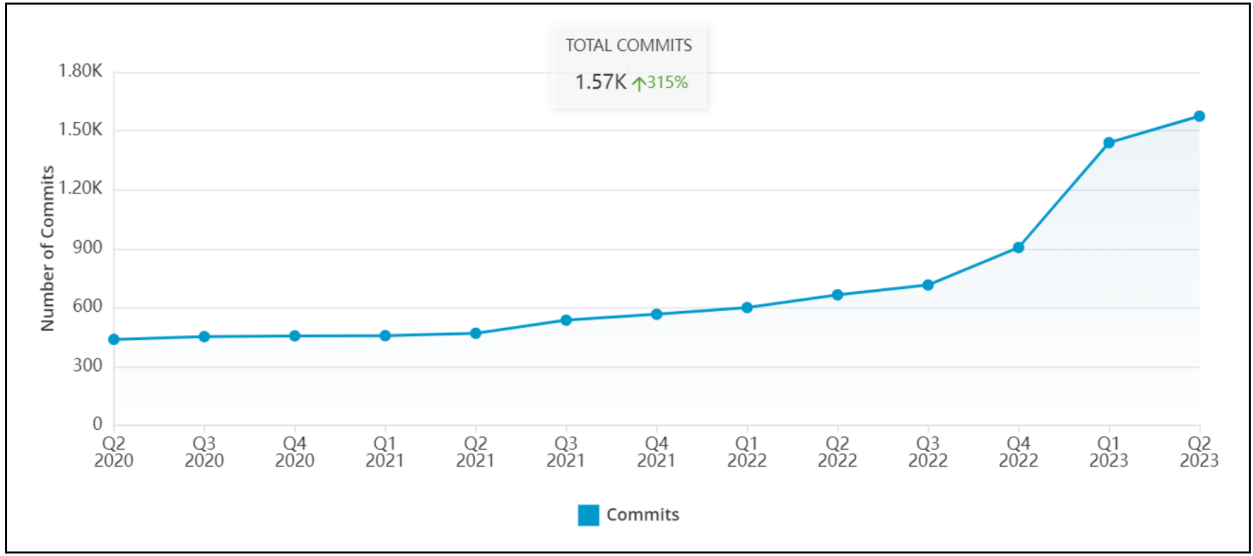

*Figure 9: Commits made across all repositories*

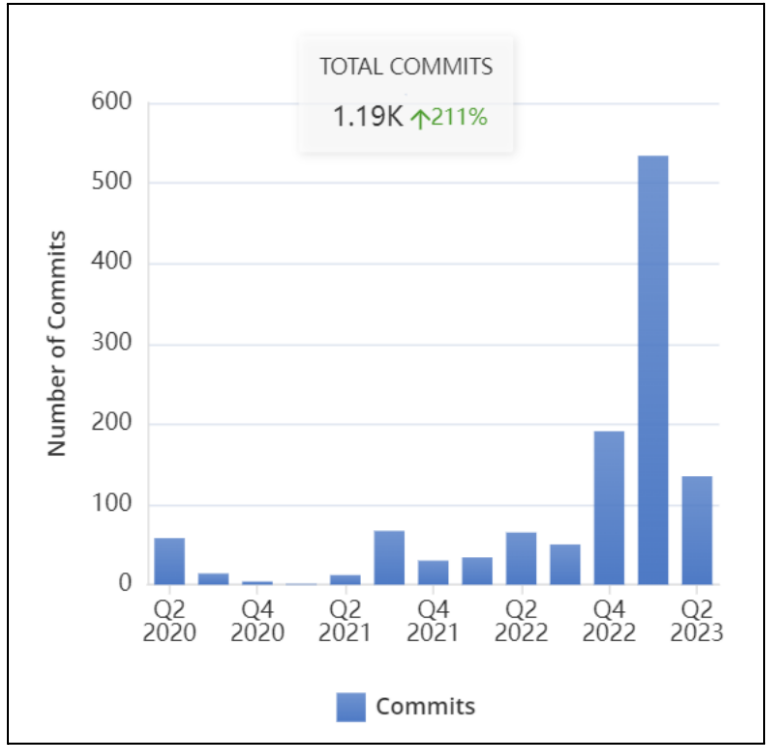

*Figure 10: Commits trend - the number of commits performed during the selected time period*

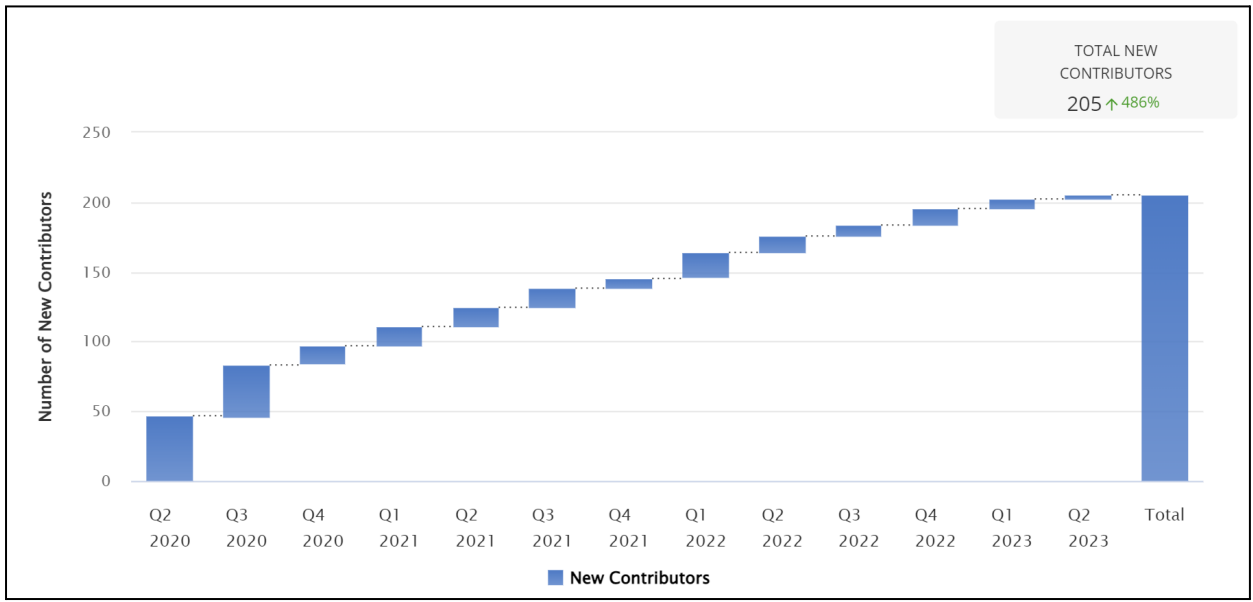

*Figure 11: Contributor growth - new contributors actively contributing to the project*

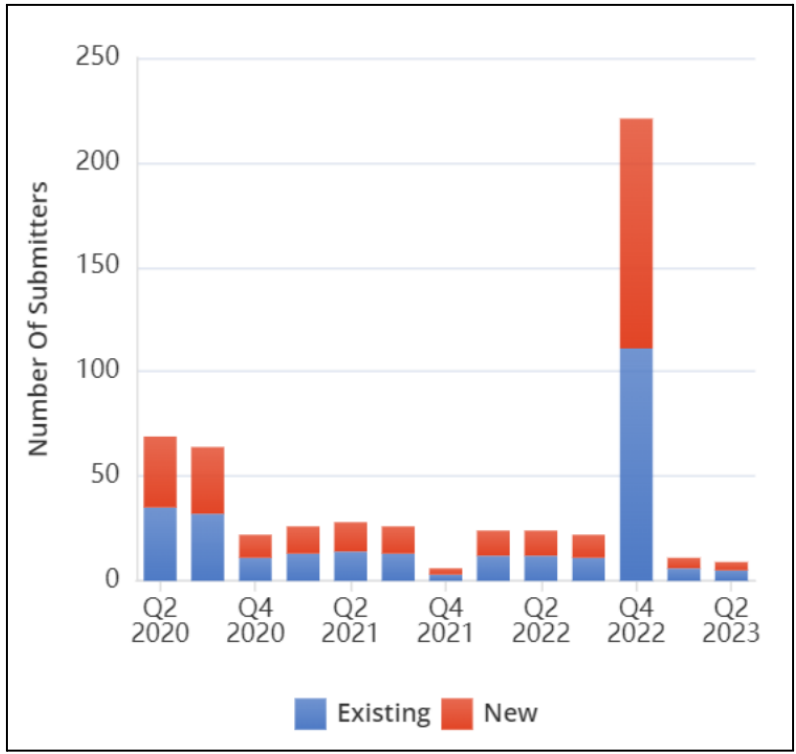

*Figure 12: Active submitters - the number of unique pull requests / changeset sumbitters showing new and repeat submitters*

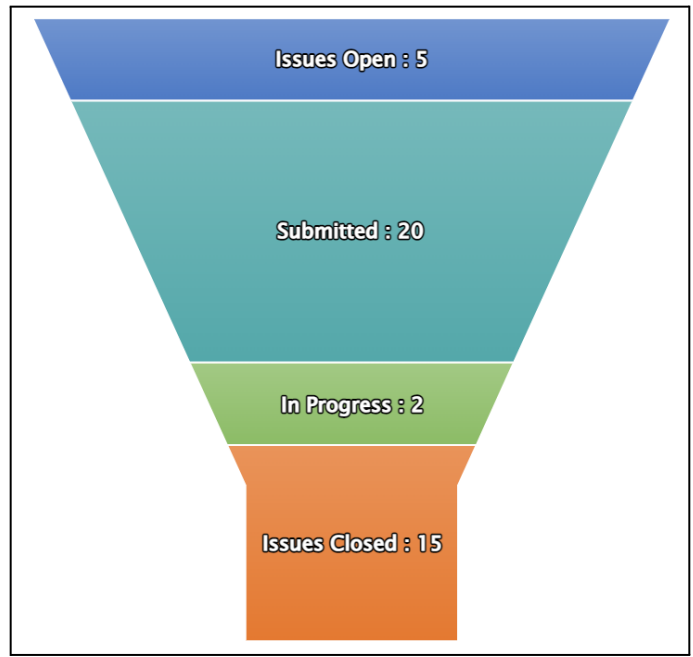

*Figure 13: Resolution pipeline - the resolution pipeline key data points such as issues waiting to be closed, pushed pull requests and more*

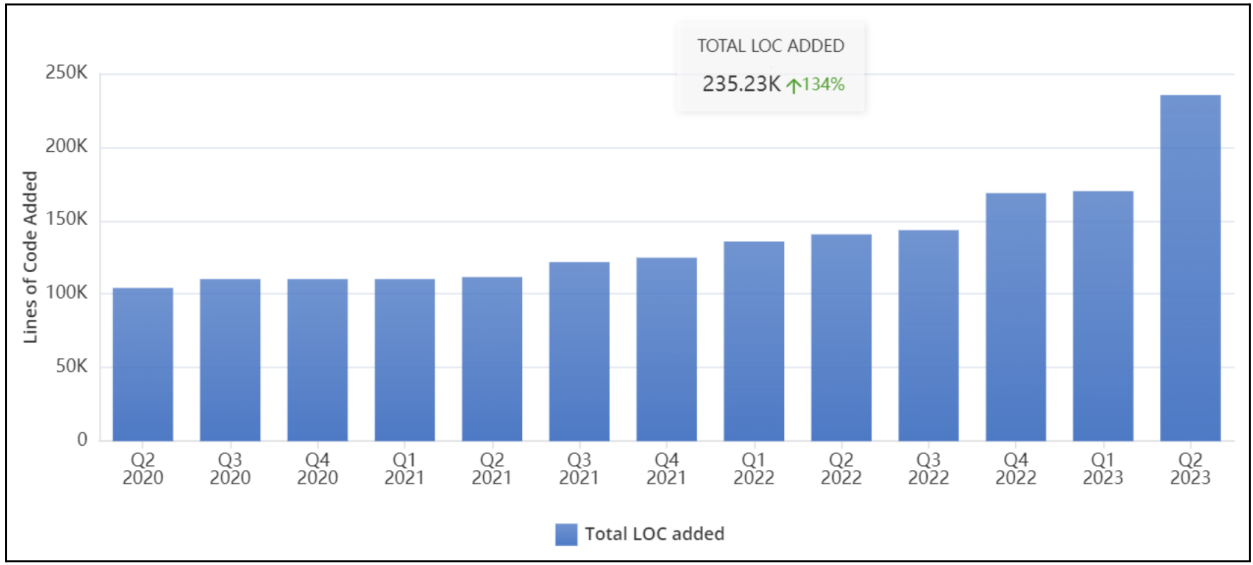

*Figure 14: Lines of code added across all unique commits*

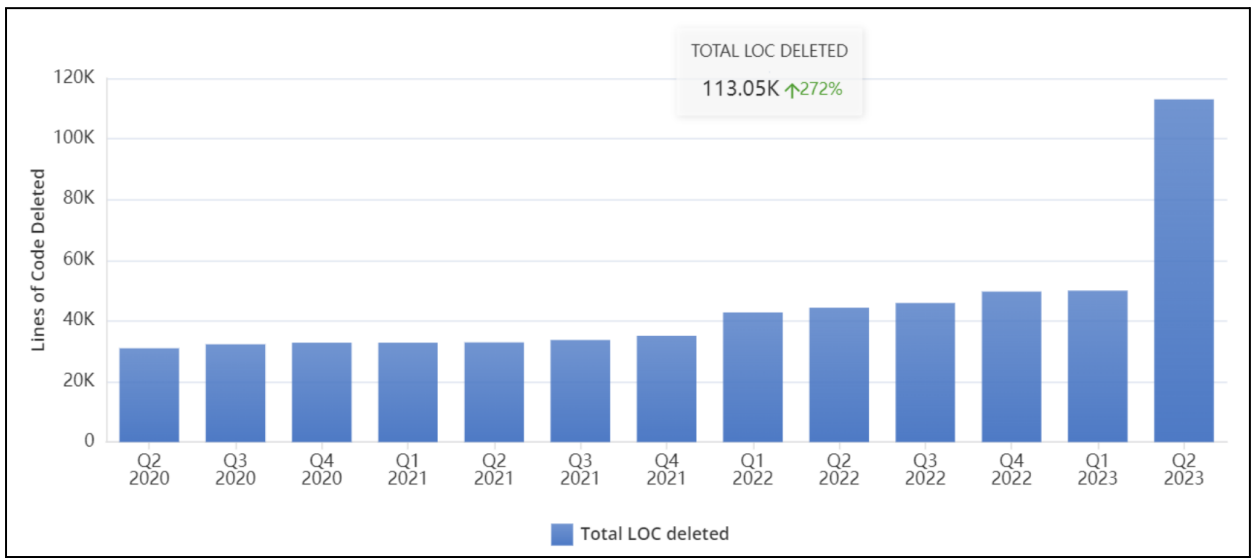

*Figure 15: Lines of code removed across all unique commits*

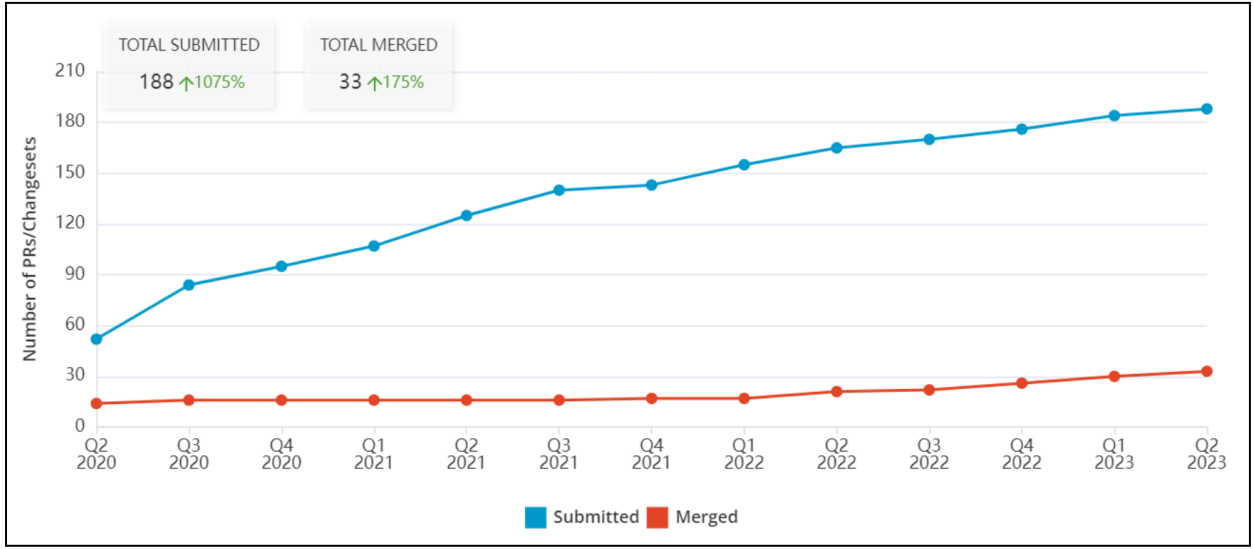

*Figure 16: Pull requests/changeset history - the total number of pull requests/changeset history submitted and merged*

# Future Work

C³ is doing something similar to what RedHat calls source-to-image (S2I), which is part of the OpenShift Source-to-Image project.

OpenShift is a container platform developed by Red Hat, and Source-to-Image (S2I) is a component of OpenShift that enables developers to build reproducible container images from source code. S2I provides a framework for creating container images that include the application source code and its dependencies, along with a builder image that defines the build process.

The OpenShift Source-to-Image (S2I) project is open source and available on GitHub. The project's repository can be found at: https://github.com/openshift/source-to-image

With the strategic direction of RedHat to focus on open source Data & AI it would be beneficial to CLAIMED and RedHat to integrate C³ into OpenShift S2I.

# Acknowledgements

The authors are deeply thankful to Kommy Weldemariam, Thomas Brunschwiler, Johannes Jakubik, Marcus Freitag, Daiki Kimura and Naomi Simumba for their valuable feedback.

## Conclusion

The concept of write once run anywhere (WORA) [wora] goes back to 1995 and the Java Virtual Machine (JVM), which was proposed as the new 'operating system' for applications of all kinds. At least since the rise of non JVM languages like Go, JavaScript and Python on the one hand and docker containerization on the other hand the WORA ecosystem slowly emerged into kubernetes being the leading container orchestrator technology and RedHat OpenShift the leading kubernetes distribution. CLAIMED builds on top of the WORA promise by creating operators once and having C³ - an implementation of RedHat's S2I concept - bringing the operators to any container platform and workflow orchestration system. As CLAIMED operators are first class citizens on Data & AI workflows, they bring speed, reproducibility, verifiability and reusability to open source science and accelerated discovery.

## References

| | |
|---|---|
| [docker] | https://en.m.wikipedia.org/wiki/Docker_(software) |
| [podman] | https://podman.io/ |
| [covidata] | https://arxiv.org/abs/2006.11988 |
| [elyra] | http://elyra.readthedocs.io/ |
| [aix360] | https://github.com/Trusted-AI/AIX360 |
| [sunsyn] | https://en.m.wikipedia.org/wiki/Sun-synchronous_orbit |
| [ibmfm] | https://research.ibm.com/topics/foundation-models |
| [rhosds] | https://developers.redhat.com/products/red-hat-openshift-data-science/ |
| [tl] | https://en.wikipedia.org/wiki/Transfer_learning |
| [sentinel2] | https://en.wikipedia.org/wiki/Sentinel-2 |
| [landsat] | https://en.wikipedia.org/wiki/Landsat_9 |
| [crs] | https://en.wikipedia.org/wiki/Spatial_reference_system |
| [osm] | https://en.wikipedia.org/wiki/OpenStreetMap |
| [watsonxai] | https://www.ibm.com/products/watsonx-ai |
| [modelmesh] | https://github.com/kserve/modelmesh |
| [wora] | https://en.wikipedia.org/wiki/Write_once,_run_anywhere |
| [mcad] | https://github.com/project-codeflare/multi-cluster-app-dispatcher |
| [claimed] | https://github.com/claimed-framework/component-library |